\def\year{2022}\relax
\documentclass[letterpaper]{article} 
\usepackage{aaai22}  
\usepackage{times}  
\usepackage{helvet}  
\usepackage{courier}  
\usepackage[hyphens]{url}  
\usepackage{graphicx} 
\usepackage{natbib}  
\usepackage{caption} 
\DeclareCaptionStyle{ruled}{labelfont=normalfont,labelsep=colon,strut=off} 
\usepackage{algorithm}
\usepackage{algorithmic}
\usepackage{amssymb}
\usepackage{amsmath}
\usepackage{mathtools}
\usepackage{multirow}
\usepackage{booktabs}
\usepackage{tablefootnote}
\usepackage{breakurl}
\usepackage{newfloat}
\usepackage{listings}
\floatstyle{ruled}
\newfloat{listing}{tb}{lst}{}
\floatname{listing}{Listing}
\title{How Does Knowledge Graph Embedding Extrapolate\\ to Unseen Data: A Semantic Evidence View}
\author{
    Ren Li,\textsuperscript{\rm 1, 2} 
    Yanan Cao,\textsuperscript{\rm 1, 2} 
    Qiannan Zhu,\textsuperscript{\rm 3, 4} 
    Guanqun Bi,\textsuperscript{\rm 1, 2}
    Fang Fang,\textsuperscript{\rm 1, 2}\thanks{Corresponding author}
    Yi Liu,\textsuperscript{\rm 5}
    Qian Li \textsuperscript{\rm 6}
}
\affiliations{
    \textsuperscript{\rm 1} Institute of Information Engineering, Chinese Academy of Sciences \\
    \textsuperscript{\rm 2} School of Cyber Security, University of Chinese Academy of Sciences \\
    \textsuperscript{\rm 3} Gaoling School of Artificial Intelligence, Renmin University of China \\
    \textsuperscript{\rm 4} Beijing Key Laboratory of Big Data Management and Analysis Methods \\
    \textsuperscript{\rm 5} National Computer Network Emergency Response Technical Team/Coordination Center of China \\
    \textsuperscript{\rm 6} University of Technology Sydney \\
    \{liren, caoyanan, biguanqun, fangfang0703\}@iie.ac.cn,
    \\zhuqiannan@ruc.edu.cn, liuyi@cert.org.cn, Qian.Li@uts.edu.au
}

\begin{document}

\maketitle

\begin{abstract}
Knowledge Graph Embedding (KGE) aims to learn representations for entities and relations. 
Most KGE models have gained great success, especially on \textbf{extrapolation} scenarios. Specifically, given an \textbf{unseen} triple $(h, r, t)$, a trained model can still correctly predict $t$ from $(h, r, ?)$, or $h$ from $(?, r, t)$, such extrapolation ability is impressive.
However, most existing KGE works focus on the design of delicate triple modeling function, which mainly tells us how to measure the plausibility of observed triples, but offers limited explanation of why the methods can extrapolate to unseen data, and what are the important factors to help KGE extrapolate. 
Therefore in this work, we attempt to study the KGE extrapolation of two problems: 1. How does KGE extrapolate to unseen data? 2. How to design the KGE model with better extrapolation ability? 
For the problem 1, we first discuss the impact factors for extrapolation and from relation, entity and triple level respectively, propose three \textbf{Semantic Evidence}s (\textbf{SE}s), which can be observed from train set and provide important semantic information for extrapolation. Then we verify the effectiveness of SEs through extensive experiments on several typical KGE methods.
For the problem 2, to make better use of the three levels of SE, we propose a novel GNN-based KGE model, called \textbf{S}emantic \textbf{E}vidence aware \textbf{G}raph \textbf{N}eural \textbf{N}etwork (\textbf{SE-GNN}). In SE-GNN, each level of SE is modeled explicitly by the corresponding neighbor pattern, and merged sufficiently by the multi-layer aggregation, which contributes to obtaining more extrapolative knowledge representation. 
Finally, through extensive experiments on FB15k-237 and WN18RR datasets, we show that SE-GNN achieves state-of-the-art performance on Knowledge Graph Completion task and performs a better extrapolation ability. Our code is available at \burl{https://github.com/renli1024/SE-GNN}.
\end{abstract}

\section{Introduction}
Knowledge Graphs (KGs) like Freebase \cite{SIGMOD_2008_Bollacker_Freebase} and WordNet \cite{ACM_1995_Miller_WordNet} are significant resources to support numerous artificial intelligence applications, such as recommendation system \cite{CIKM_2018_Wang_RippleNet},  question answering \cite{Yasunaga_NAACL_2021_QA-GNN} and text generation \cite{ACL_2020_Zhang_Generation}, etc. KGs store graph-structured knowledge in triple form $(h, r, t)$. To integrate symbolic knowledge into numerical down-stream applications, Knowledge Graph Embedding (KGE) technique that attempts to encode the relations and entities into low-dimensional embeddings, has attracted increasing attention. 
The core idea of KGE is to design triple modeling function $f(h, r, t)$, that can predict correct tail entity $t$ from $(h, r, ?)$, or head entity $h$ from $(?, r, t)$, by scoring high for positive triple $(h, r, t)$, and low for negative triples $(h', r, t)$ and $(h, r, t')$\footnote{This prediction process is also called Knowledge Graph Completion task, which shares many common concepts with Knowledge Graph Embedding.}.

Many KGE models have been proposed and can be categorized into three families \cite{TKDE_2017_Wang_Survey, 2020_Arora_KG-GNN}, which are Translational Distance Models like TransE \cite{NeurIPS_2013_Bordes_TransE}, RotatE \cite{ICLR_2019_Sun_RotatE}; Semantic Matching Models like DistMult \cite{ICLR_2015_Yang_DistMult}, ComplEx \cite{ICML_2016_Trouillon_ComplEx}, ConvE \cite{AAAI_2018_Dettmers_ConvE_WN18RR}; and GNN-based Models like R-GCN \cite{ESWC_2018_Schlichtkrull_R-GCN}, CompGCN \cite{ICLR_2020_Vashishth_CompGCN}.  
Most of these KGE models have gained great success, especially on \textbf{extrapolation} scenarios, which is that given an \textbf{unseen} triple $(h, r, t)$, a well trained model can still correctly predict $t$ from $(h, r, ?)$ or $h$ from $(?, r, t)$, such ability is impressive.
However, most existing KGE works focus on the design of delicate triple modeling function, but explains little about why the methods can extrapolate to unseen data, and what are the important factors to help KGE extrapolate. Therefore in this work, we attempt to, from a data relevant view, study KGE extrapolation of two problems: 1. How does KGE extrapolate to unseen data? 2. How to design the KGE model with better extrapolation ability? 

\begin{figure}[t]
    \centering 
    \includegraphics[width=\columnwidth]{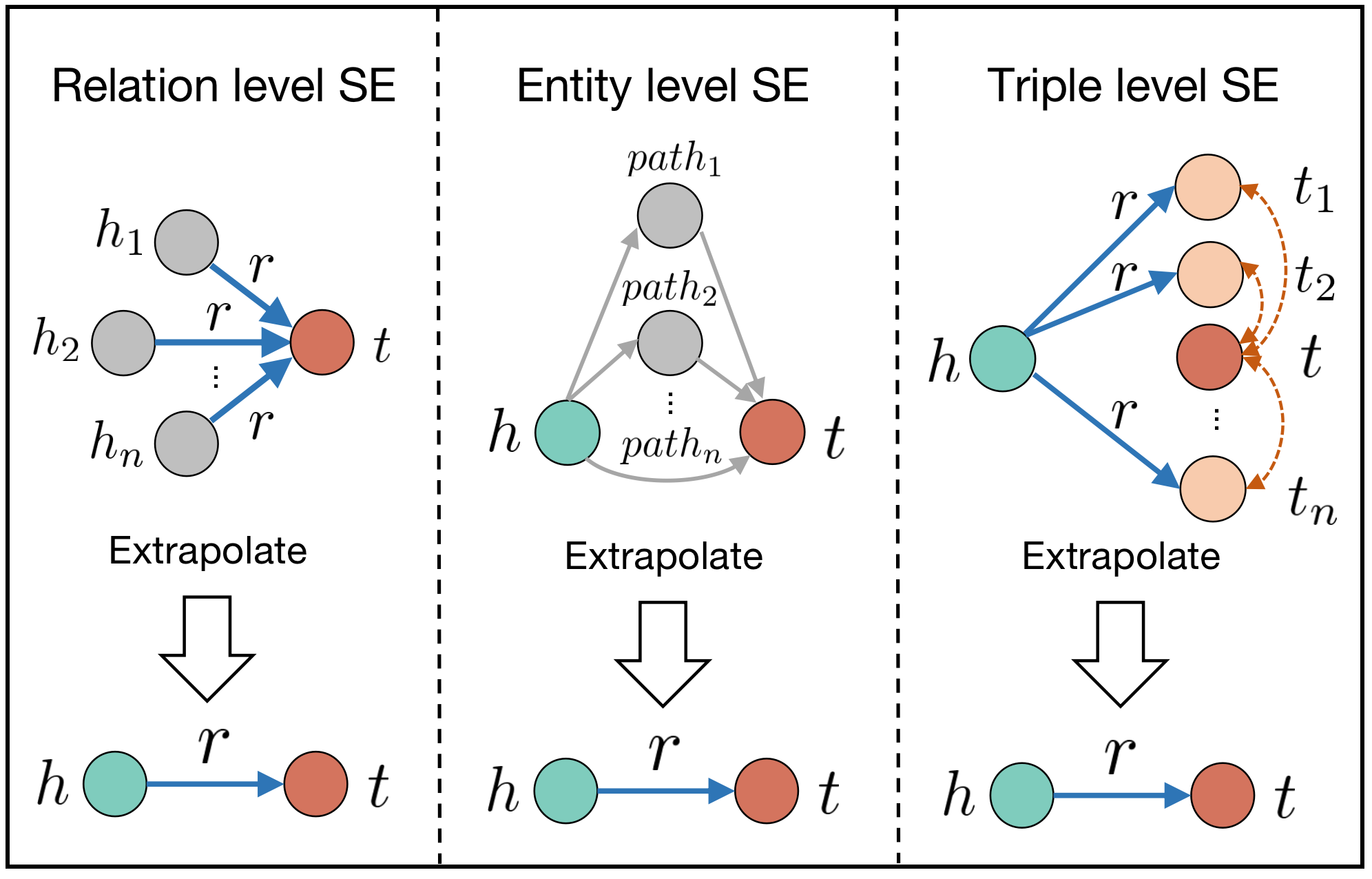}
    \caption{The demonstration of three levels of Semantic Evidence.} 
    \label{fig: SE} 
\end{figure}

For the problem 1, for an unseen triple $(h, r, t)$, we treat the prediction from $(h, r, ?)$ to $t$ with a semantic matching idea. For a good extrapolative matching, $(h, r, ?)$ and $t$ must have obtained some semantic relatedness during training, and we consider the relatedness may come from three levels: the individual $r$ part with $t$ (relation level), the individual $h$ part with $t$ (entity level), and the combination $(h, r, ?)$ part with $t$ (triple level). 
We name such three factors as \textbf{Semantic Evidence (SE)}, to indicate the supporting semantic information they provide for extrapolation. Then, we quantify the SEs with three corresponding metrics respectively. 
For relation level, it is measured by the co-occurrence of $r$ and $t$ in train set; for entity level, it is the path connections from $h$ to $t$ in train set; for triple level, it is the similarity between existed ground truth entities of $(h, r, ?)$ and $t$. The demonstration of three levels of SE can be seen in figure \ref{fig: SE}. 
Furthermore, we verify the effectiveness of SEs through extensive experiments on several typical KGE methods, and demonstrate that SEs serve as an important role for understanding the extrapolation ability of KGE.

For the problem 2, based on the conclusion of problem 1, Semantic Evidences are important for designing KGE models with powerful extrapolation ability. However, current works capture the SE information mainly through an implicit and insufficient way, which limits their extrapolation performance. 
Hence in this work, to make better use of the three levels of SE, we propose a novel GNN-based KGE model, called \textbf{S}emantic \textbf{E}vidence aware \textbf{G}raph \textbf{N}eural \textbf{N}etwork (\textbf{SE-GNN}). In SE-GNN, each level of SE is modeled explicitly by corresponding neighbor pattern, and merged sufficiently by the multi-layer aggregation mechanism of GNNs, which contributes to obtaining more extrapolative knowledge representation. The model architecture is demonstrated in figure \ref{fig: model}.

In summary, our main contributions are as follows: 
\begin{itemize}
    \item We are the first to explore KGE extrapolation problem, from a data relevant and model independent view, and further introduce three levels of Semantic Evidence to understand the extrapolation ability of KGE. We also conduct extensive experiments on various typical KGE models to verify our assumption. 
    \item We dive into the way of designing the KGE model with better extrapolation ability, through explicitly and sufficiently modeling the Semantic Evidences into knowledge embedding. We propose a novel GNN-based KGE method called SE-GNN, which helps the learned knowledge representation achieve more improved extrapolation performance. 
    \item Through extensive experiments on FB15k-237 and WN18RR datasets of Knowledge Graph Completion task, we demonstrate the validity of our introduced Semantic Evidence concept and SE-GNN method.
\end{itemize}

\section{Related Work}

\paragraph{Knowledge Graph Embedding}
Knowledge Graph Embedding is an active research area. Based on the scoring function and whether global graph structure is utilized, literature works can be divided into three families \cite{TKDE_2017_Wang_Survey, 2020_Arora_KG-GNN}.
(\romannumeral1) \textbf{Translational Distance Models} apply distance-based scoring functions and model relations as some operation, like addition operation in TransE \cite{NeurIPS_2013_Bordes_TransE}, hyper-plane addition in TransH \cite{AAAI_2014_Wang_TransH}, 
complex field rotation in RotatE \cite{ICLR_2019_Sun_RotatE}, etc. 
(\romannumeral2) \textbf{Semantic Matching Models} utilize similarity-based scoring function. DistMult \cite{ICLR_2015_Yang_DistMult} proposes a multiplication model to represent the likelihood of a fact. ComplEx \cite{ICML_2016_Trouillon_ComplEx} models the triple matching function in complex domain. ConvE \cite{AAAI_2018_Dettmers_ConvE_WN18RR}, InteractE \cite{AAAI_2020_Vashishth_InteractE} apply neural network for similarity modeling. 
(\romannumeral3) \textbf{GNN-based Models} tend to capture the structure characteristics of KGs through Graph Neural Networks. R-GCN \cite{ESWC_2018_Schlichtkrull_R-GCN} introduces a relation-specific transformation to merge the relation information when neighbor aggregating. CompGCN \cite{ICLR_2020_Vashishth_CompGCN} proposes various composition operations for neighbor aggregation to model the structure pattern of multi-relational graph. 

\paragraph{Extrapolation Ability Study}
In Machine Learning Theory field, there are many works that attempt to study the generalization and extrapolation ability of Neural Networks or Multilayer perceptrons (MLPs) \cite{IJCNN_1992_Haley_Extrapolation, CSM_1992_Barnard_Extrapolation, NeurIPS_2019_Bietti_Inductive, ICLR_2020_Ba_Generalization, ICLR_2021_Xu_NNExtrapolate}. Like in \cite{ICLR_2021_Xu_NNExtrapolate}, it is proved that ReLU MLPs can not extrapolate most nonlinear functions, but can extrapolate linear function when the training distribution is sufficiently diverse. And for Graph Neural Networks, it is showed that they can encode non-linearity in architecture and features to help extrapolation. However, the conclusions of above works cannot directly apply to KGE field. Because the analysis of Neural Networks mostly concentrates on classification or regression task, with only one single object or distribution. For Graph Neural Networks, the study is also mainly about node classification or graph classification task. While for KGE task, there are three targets $(h, r, t)$ mutually influencing and serving as a matching task between $(h, r, ?)$ and $t$, which makes the extrapolation analysis of KGE differs from the correspondence in ML field. In addition, in Knowledge Graphs there are abundant data pattern and fact interdependency that can be mined, which is very important to understand the extrapolation performance of KGE. Therefore, in this work we focus on a data relevant and model independent view to study the KGE extrapolation problem. 

\section{Knowledge Graph Embedding Extrapolation}
\label{sec: KGE Extrapolation}
In this section, we firstly give the definition of KGE extrapolation problem. Then we introduce three levels of Semantic Evidence to explain the extrapolation ability of KGE models. Finally we conduct experiments on various typical KGE models to verify our assumption.

\subsection{Problem Definition}
A knowledge graph is denoted as $\mathcal{G}=(\mathcal{E}, \mathcal{R}, \mathcal{F})$, where $\mathcal{E}$ and $\mathcal{R}$ represent the set of entities and relations, and $\mathcal{F} = \{(h, r, t)\} \subseteq \mathcal{E} \times \mathcal{R} \times \mathcal{E}$ is the set of triple facts. For the KGE learning process, firstly $\mathcal{F}$ will be partitioned into train, valid and test set, denoted as $\mathcal{F}_{tr}$, $\mathcal{F}_{va}$, $\mathcal{F}_{te}$ respectively. The model will be trained on $\mathcal{F}_{tr}$ and the best parameters will be selected according to $\mathcal{F}_{va}$, then the extrapolation performance will be evaluated on unseen dataset $\mathcal{F}_{te}$. 

KGE task aims to predict $t$ given $(h, r, ?)$, or $h$ given $(?, r, t)$. 
Here we treat the prediction task with the idea of semantic matching between query and answer, and without loss of generality, we denote both directions as $\mathrm{query}(h, r) \rightarrow t$. 
Under such denotation, the extrapolation problem we want to study is that why the KGE model is only trained for high scoring of $\mathrm{query}(h, r) \rightarrow t, (h, r, t) \in \mathcal{F}_{tr}$, but can still perform well for unseen data $\mathrm{query}(h, r) \rightarrow t, (h, r, t) \in \mathcal{F}_{te}$\footnote{The unseen data does not mean the new entity or relation, but the new triple combination. In fact all the entities and relations in $\mathcal{F}_{te}$ should occur in $\mathcal{F}_{tr}$, in order to learn their embeddings.}.

\subsection{Extrapolate with Semantic Evidences}
\label{sec: SE}
For a good extrapolative matching $\mathrm{query}(h, r) \rightarrow t$, $\mathrm{query}(h, r)$ and $t$ must have obtained some semantic relatedness during training. We consider the relatedness may come from three levels: the individual $r$ part with $t$ (relation level), the individual $h$ part with $t$ (entity level) and the combination $\mathrm{query}(h, r)$ part with $t$ (triple level), demonstrated as follows: 
\begin{itemize}
    \item \textbf{Relation level} relatedness between $r$ and $t$: In train set if $t$ frequently occur with queries containing $r$, i.e. there are many $\mathrm{query}(h_i, r) \rightarrow t$ in $\mathcal{F}_{tr}$, the $r$ will contain information to predict $t$. From intuition this can be regarded as entity type information. Instantly, for $\mathrm{query}(h_i, born\_in)$, the probability of predicting location $Florida$ should be higher than predicting movie $Iron\_Man$, no matter what the specific $h_i$ is.
    \item \textbf{Entity level} relatedness between $h$ and $t$: In train set if there are observed queries or indirect queries from $h$ to $t$, this will close their semantic relevancy and provide evidences for other queries between $h$ and $t$. 
    For example, $\mathrm{query}(h, is\_mother) \rightarrow e_1$ and $\mathrm{query}(e_1, is\_father) \rightarrow t$ will bring confidence for predicting $\mathrm{query}(h, is\_grandmother) \rightarrow t$. Under the graph view, this can be regarded as the \textit{path} from $h$ to $t$. 
    \item \textbf{Triple level} relatedness between $\mathrm{query}(h, r)$ and $t$: For $\mathrm{query}(h, r)$, it may exist other ground truth entities $t'$ in train set. If the model has been trained for $\mathrm{query}(h, r) \rightarrow t'$, meanwhile $t$ and $t'$ share much similarity, it will be natural for the model to extrapolate to $\mathrm{query}(h, r) \rightarrow t$. For example, if we have known \texttt{query(James\_Cameron, profession)} $\rightarrow$ \texttt{film\_director} and \texttt{screen\_writer}, it is not difficult to predict \texttt{query(James\_Cameron, profession)} $\rightarrow$ \texttt{film\_producer}.
\end{itemize}

All above relatednesses are from train set and can be observed, so for a KGE model, though it does not train for the unseen data $\mathrm{query}(h, r) \rightarrow t$, it has gained enough information from observed triples to make the prediction. We name such relatedness as \textbf{Semantic Evidence (SE)}, to indicate the supporting semantic information they provide for extrapolation. We demonstrate the three levels of SE in figure \ref{fig: SE}. In addition, we also do the extensive case study for the three levels of SE, to provide an intuitive demonstration about how the Semantic Evidence helps extrapolate. The case study content is placed in appendix \ref{ap: case study} because of the space limitation.

\subsection{Experiment Verification}
\begin{figure*}[t]
    \centering
    \includegraphics[width=\textwidth]{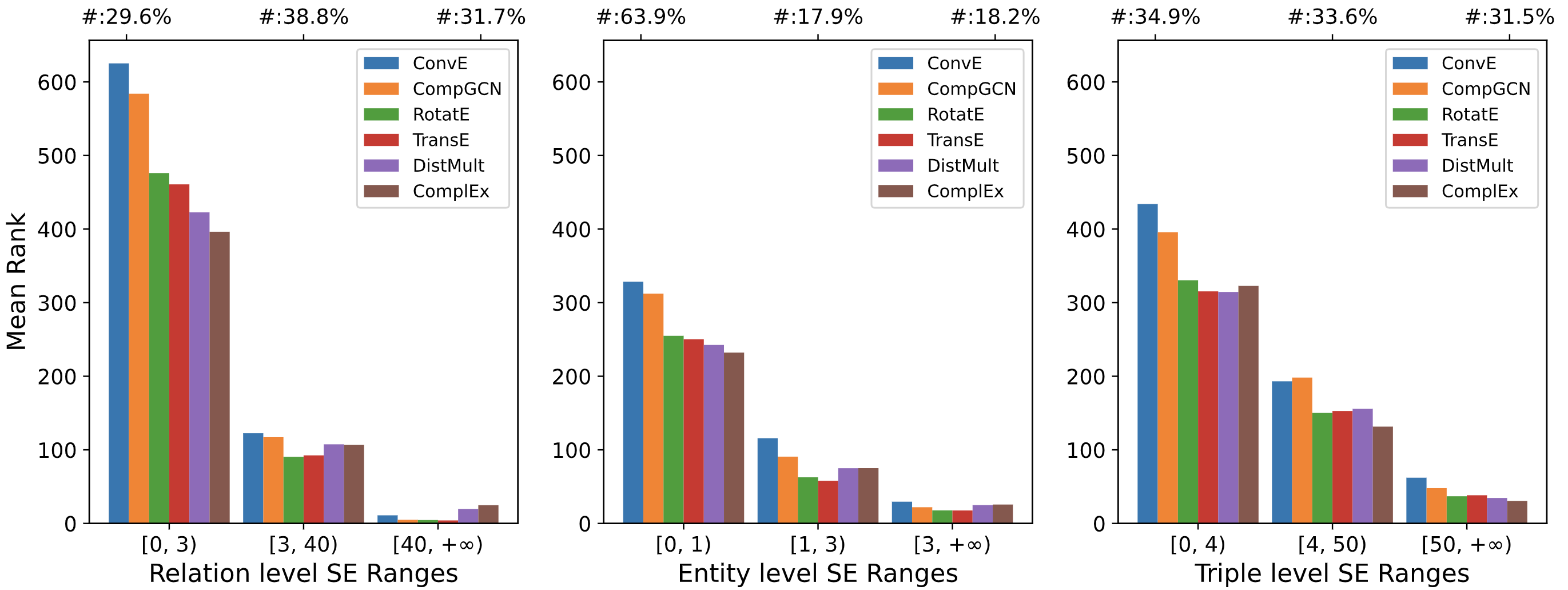}
    \caption{
        KGE extrapolation performance on different SE ranges. The test data of FB15k-237 is divided into three ranges based on evidence metric. Bottom x-axis denotes the metric value, top x-axis denotes the data portion of each range and y-axis denotes the Mean Rank of model prediction result 
        (low value indicates good performance, and $1$ is the best).
    } 
    \label{fig: fb_baselines_se2rank} 
\end{figure*}

\label{sec: SE Verification}
In this section, we attempt to verify the effectiveness of the proposed SE concept through experiments. Firstly, for an unseen prediction $\mathrm{query}(h, r) \rightarrow t$, we propose three corresponding metrics to quantify the evidence strength of each SE as follows: 

\begin{itemize}
    \item $S_{rel}$ for relation level SE: It is the number of triples in train set that satisfy $(h_i, r, t)$, which can be formulated as: 
    $$
    S_{rel}=|\{(h_i, r, t) | (h_i, r, t) \in \mathcal{F}_{tr}\}|
    $$ 
    where $|\mathrm{set}|$ denotes the element number of a set. 
    \item $S_{ent}$ for entity level SE: It is the number of path from $h$ to $t$ in train set, indicating the semantic relevancy of $h$ and $t$. For simplification, we limit the path length $\leq 2$. $S_{ent}$ is formulated as:
    \begin{align*}
        S_{ent} =& |\{(h, r_i, t) | (h, r_i, t) \in \mathcal{F}_{tr}\}| + \\
        & |\{(h, r_i, e_k, r_j, t) | (h, r_i, e_k), (e_k, r_j, t) \in \mathcal{F}_{tr}\}|
    \end{align*} 
    \item $S_{tri}$ for triple level SE: It is the similarity measurement between $t$ and $\mathrm{query}(h, r)$'s ground truth entity $t'$ in train set:  
    $$
    S_{tri} = \sum_{t'} \mathrm{Sim}(t, t'), \, (h, r, t') \in \mathcal{F}_{tr}
    $$
    For similarity function $\mathrm{Sim}(t, t')$, though there have been proposed many entity similarity computing methods for KGs \cite{IKDD_2015_Choudhury_SimCat, TKDE_2017_Zhu_Concept, IEEE_2018_Sun_Topic_Sim}, most of them need external information like entity category, description text, etc. Here we hope for a method that only relates to KG itself, so we take the idea of Distributional Semantic Hypothesis: \textit{words that are used and occur in the same contexts tend to purport similar meanings} \cite{1954_Harris_Distribution}, and measure the entity similarity according to its neighbor structure (context). 
    This can be formulated as the number of common neighbor entity-relation pairs that $t$ and $t'$ share:
    \begin{align*}
        \mathrm{Sim}(t, t') = & \left| \{(h_i, r_i)| (h_i, r_i, t) \in \mathcal{F}_{tr}\} \, \cap \right. \\
        & \left. \{(h_i, r_i)| (h_i, r_i, t') \in \mathcal{F}_{tr}\} \right|
    \end{align*}
\end{itemize}

Then we reproduce several typical KGE models and analyze their extrapolation performance under different SE configurations. Specifically, we use FB15k-237 dataset \cite{2015_Toutanova_FB15k-237}, a frequently used public KG dataset, and compute the above three SE metrics for each data $\mathrm{query}(h, r) \rightarrow t$ in test set. For each SE, we divide the data evenly into three ranges with ascending order of metric value. Hence the three ranges represent the low, medium and high evidence strength respectively. 
One exception is for entity level SE, that because the range [0, 1) cannot be divided further, the proportion of three ranges is about 6:2:2.
Then we select six typical KGE models of different types, which are TransE, RotatE (Translational Distance Models), DistMult, ComplEx, ConvE (Semantic Matching Models), CompGCN (GNN-based Models), and evaluate their prediction results on each SE range. The results are demonstrated in figure \ref{fig: fb_baselines_se2rank}. 

We can see that for \textbf{all models} it exists a consistent better prediction result with evidence strength increasing. When there is abundant SE, all the KGE models can perform a good extrapolation result, like the rightmost range of each SE in figure \ref{fig: fb_baselines_se2rank}. And if the SE is lacked, the models' extrapolation ability will also be limited, like the leftmost range. 
In addition, we also conduct the similar experiment verification on WN18RR dataset \cite{AAAI_2018_Dettmers_ConvE_WN18RR}, and the results are placed in figure \ref{fig: wn_baselines_se2rank} of appendix \ref{ap: wn_baselines_se2rank}. It can be seen that there is the same phenomenon on WN18RR dataset. 
This further verifies the strong correlation between SE and extrapolation performance. That is to say, regardless of the specific method selected, the models always extrapolate well to data with high SE evidence, which verifies that the proposed SE is a reasonable data view explanation to understand the impressive extrapolation ability of KGE.

\section{Semantic Evidence aware GNN}
In this section, to make better use of the Semantic Evidence information for more extrapolative knowledge representation, we propose a novel GNN-based KGE model called \textbf{S}emantic \textbf{E}vidence aware \textbf{G}raph \textbf{N}eural \textbf{N}etwork (\textbf{SE-GNN}), which is designed to model the three levels of SE explicitly and sufficiently. 

\begin{figure}[t]
    \centering 
    \includegraphics[width=\columnwidth]{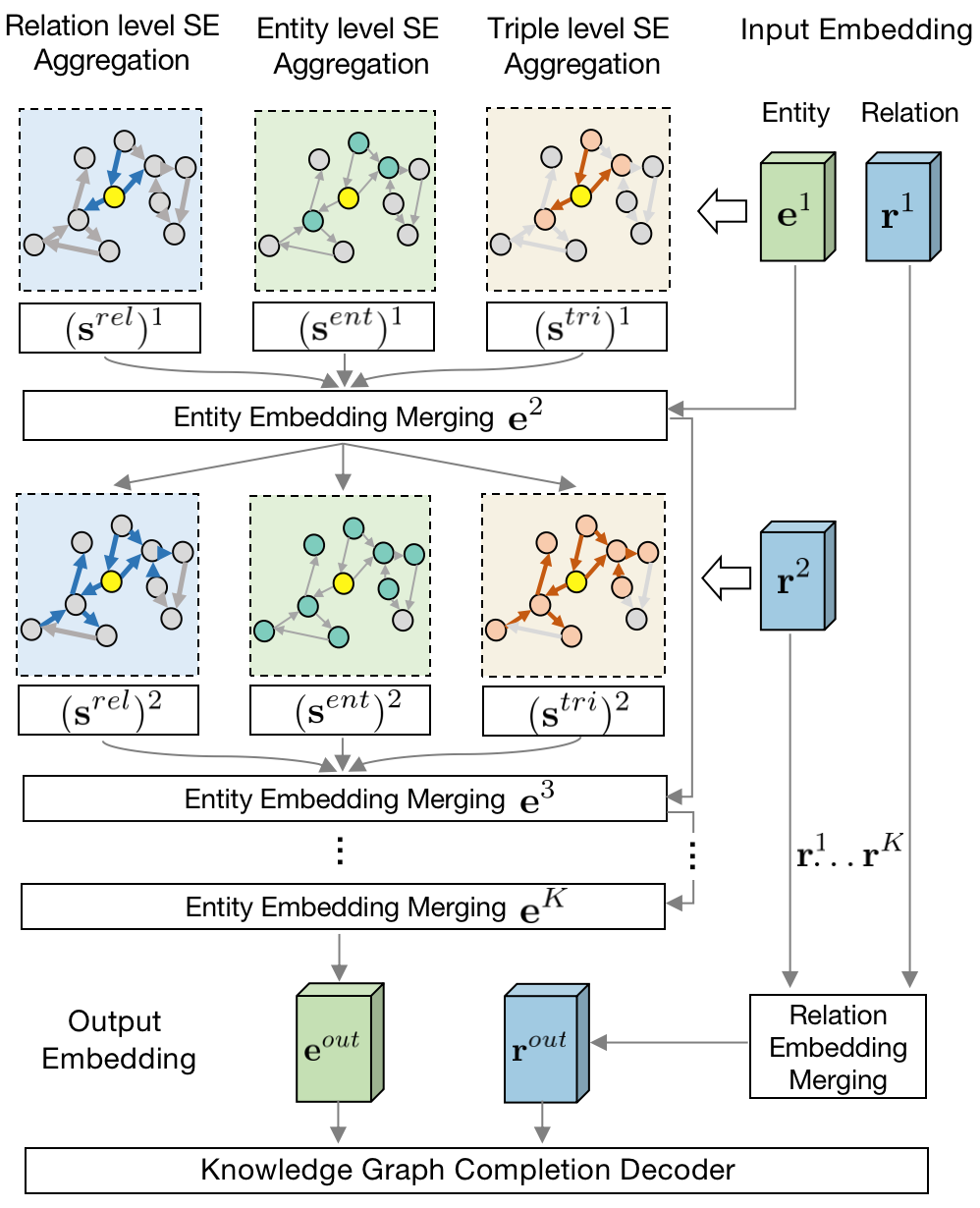}
    \caption{The model architecture of SE-GNN. The blue, green and orange graph represents relation, entity and triple level SE aggregation process respectively. The yellow node is the example center node of neighbor aggregation. 
    By layer-wise iteration, SE-GNN can access a wide range of graph structure and model the deep interaction of SEs. 
    Finally, the output entity and relation embedding are fed into a Knowledge Graph Completion decoder to perform the extrapolation. } 
    \label{fig: model}
\end{figure}

\subsection{Modeling SEs with Neighbor Pattern}
\label{sec: modeling_se}
Knowing from previous section, Semantic Evidences are important to design KGE models with powerful extrapolation ability. 
However, for most current KGE works, there is no awareness of such extrapolation factors and they capture the SE information mainly through an implicit and insufficient way, which limits their extrapolation performance.
Hence in this work, we explicitly treat each SE as different neighbor pattern and model them sufficiently through multi-layer aggregation mechanism of GNNs, for obtaining more extrapolative knowledge representations.

Specifically, for relation level SE, it describes the overall relation-entity interactions, which can be captured through neighbor relation pattern of an entity. By aggregating all the connected relations, we can get the representation as: 
\begin{equation}
    \mathbf{s}_i^{rel} = \sigma\left( \sum_{(e_j, r_j) \in \mathcal{N}_i} \alpha^{rel}_{ij}\, W^{rel} \,\mathbf{r}_j \right)
\end{equation}
$\mathbf{s}_i^{rel} \in \mathbb{R}^n$ denotes the relation level SE representation of entity $e_i$, where $n$ is hidden dimension. $\mathbf{r}_j \in \mathbb{R}^n$ is the embedding of relation $r_j$. $\mathcal{N}_i = \{(e_j, r_j)|(e_j, r_j, e_i) \in \mathcal{F}_{tr} \}$ denotes $e_i$'s neighbor entities, associated with connecting relation in train set. $W^{rel} \in \mathbb{R}^{n \times n}$ is linear transformation matrix and $\sigma$ is non-linear activation function. $\alpha^{rel}_{ij}$ is aggregation attention, which is computed as: 
\begin{equation}
    \alpha_{ij}^{rel} = \frac{\exp \left( \mathbf{r}_j^{\mathrm{T}} \, \mathbf{e}_i \right)}{\sum_{(e_k, r_k) \in \mathcal{N}_i} \exp \left( \mathbf{r}_k^{\mathrm{T}} \, \mathbf{e}_i \right)}
\end{equation}
$\mathbf{e}_i \in \mathbb{R}^n$ is the embedding of entity $e_i$. We use dot product to dynamically compute the attention importance of neighbor relation $r_j$ to center entity $e_i$. 

For entity level SE, it describes the path connection information between entities, and can be captured from neighbor entity pattern. With aggregating neighbor entities once, we can capture all the $1$ length paths, and we can access to longer paths by iterative multi-layer aggregation. Here we only introduce a single layer formulation and we will introduce the whole model architecture in section \ref{sec: model arch}: 
\begin{equation}
    \mathbf{s}_i^{ent} = \sigma\left( \sum_{(e_j, r_j) \in \mathcal{N}_i} \alpha^{ent}_{ij} \, W^{ent} \, \mathbf{e}_j \right)
\end{equation}
$\mathbf{s}_i^{ent} \in \mathbb{R}^n$ is the entity level SE representation of $e_i$. $\alpha^{ent}_{ij}$ is aggregated attention and computed as follows: 
\begin{equation}
    \alpha_{ij}^{ent} = \frac{\exp \left( \mathbf{e}_j^{\mathrm{T}} \, \mathbf{e}_i \right)}{\sum_{(e_k, r_k) \in \mathcal{N}_i} \exp \left(\mathbf{e}_k^{\mathrm{T}} \, \mathbf{e}_i \right)}
\end{equation}

For triple level SE, it describes the triple similarity characteristics from the neighbor structure view, where both neighbor entities and relations should be considered. We design the aggregation function as: 
\begin{equation}
    \label{eq: triple_se_agg}
    \mathbf{s}_i^{tri} = \sigma\left( \sum_{(e_j, r_j) \in \mathcal{N}_i} \alpha^{tri}_{ij} \, W^{tri} \, \varphi(\mathbf{e}_j, \mathbf{r}_j) \right)
\end{equation}
where $\varphi(\mathbf{e}, \mathbf{r})$ is the composition function to fuse the entity and relation information. The selection includes addition function: $\varphi(\mathbf{e}, \mathbf{r}) = \mathbf{e} + \mathbf{r}$; multiplication function: $\varphi(\mathbf{e}, \mathbf{r}) = \mathbf{e} * \mathbf{r}$; Multilayer Perceptron: $\varphi(\mathbf{e}, \mathbf{r}) = \mathrm{MLP}([\mathbf{e} || \mathbf{r}])$, where $||$ is vector concatenation operation. The attention weights $\alpha^{tri}_{ij}$ is computed similarly as: 

\begin{equation}
    \label{eq: triple_se_attention}
    \alpha_{ij}^{tri} = \frac{\exp \left( \varphi(\mathbf{e}_j, \mathbf{r}_j)^{\mathrm{T}} \, \mathbf{e}_i \right)}{\sum_{(e_k, r_k) \in \mathcal{N}_i} \exp \left( \varphi(\mathbf{e}_k, \mathbf{r}_k)^{\mathrm{T}} \,  \mathbf{e}_i\right)}
\end{equation}

\subsection{Model Architecture}
\label{sec: model arch}
In section \ref{sec: modeling_se}, we introduce the neighbor aggregation method to model each SE and obtain the corresponding representation of $\mathbf{s}_i^{rel}$, $\mathbf{s}_i^{ent}$ and $\mathbf{s}_i^{tri}$. The three embeddings provide important evidences to help the model extrapolate. We merge them with original knowledge embedding as: 
\begin{equation}
    \mathbf{e}'_i = \mathbf{e}_i + \mathbf{s}_i^{rel} + \mathbf{s}_i^{ent} + \mathbf{s}_i^{tri}
    \label{eq: 1_layer_agg}
\end{equation}
This can be seen as one single aggregation layer of GNN, which only captures the SE information in 1-hop neighborhood. To acquire the multi-hop neighbor information and model the deeper interaction of SE components, we introduce a multi-layer version for SE aggregation, which is demonstrated in figure \ref{fig: model}. We take the output $\mathbf{e}'_i$ as the next layer's input, and aggregate iteratively as:
\begin{align}
    \label{eq: n_layer_agg}
    \mathbf{e}^{l+1}_i &= \mathbf{e}^{l}_i + (\mathbf{s}_i^{rel})^{l} + (\mathbf{s}_i^{ent})^{l} + (\mathbf{s}_i^{tri})^{l}
\end{align}
$\mathbf{e}^{l+1}_i$ denotes the embedding of $e_i$ in $(l+1)$-th layer. $(\mathbf{s}_i^{tri})^{l}$ is the triple level SE embedding in $l$-th layer, which is computed from $\mathbf{e}^{l}_i$ and $\mathbf{r}^{l}_i$: 
\begin{equation}
    (\mathbf{s}_i^{tri})^{l} = \sigma\left( \sum_{(e_j, r_j) \in \mathcal{N}_i} (\alpha^{tri}_{ij})^{l} \, (W^{tri})^{l} \, \varphi(\mathbf{e}^{l}_j, \mathbf{r}^{l}_j) \right)
\end{equation}
where $(\alpha^{tri}_{ij})^{l}$ is computed in the same way as equation \ref{eq: triple_se_attention}. The layer-wise embedding of $(\mathbf{s}_i^{rel})^{l}$ and $(\mathbf{s}_i^{ent})^{l}$ can be obtained in the similar way. 
In first layer, $\mathbf{e}^{1}$ is the initialized embedding, and after $K$ layers' aggregation, we take $\mathbf{e}^{K}$ as the output entity embedding.

With regard to relation embedding, we initialize different $\mathbf{r}^{l}$ for each layer, in the consideration that relations may play a different role in different layer.  
For output relation embedding, we concat all the $\mathbf{r}^l$ used and merge them together by a transform matrix $W^{out}$. Hence the output of SE-GNN is formulated as:  
\begin{equation}
    \label{eq: SE-GNN output}
    \begin{split}
        \mathbf{e}^{out} &= \mathbf{e}^{K} \\ 
        \mathbf{r}^{out} &= W^{out} \, \mathrm{Concat}(\{\mathbf{r}^{l} | l=1,...,K\}) 
    \end{split}
\end{equation}

Then we utilize the output embedding to perform the prediction from $(h, r, ?)$ to $t$ or from $(?, r, t) $ to $h$. To align with the terminology in previous work, here we also denote this process as Knowledge Graph Completion (KGC) task. 
We choose ConvE \cite{AAAI_2018_Dettmers_ConvE_WN18RR} as our decoder, which uses 2D convolutional neural network to match $\mathrm{query}(h, r)$ and answer $t$. We refer readers to original paper for more details, and here we directly denote the model function as:
\begin{equation}
    \mathbf{q} = \mathrm{ConvE}(\mathbf{h}, \mathbf{r})
\end{equation} 
$\mathbf{q} \in \mathbb{R}^n$ is the computed query embedding. $\mathbf{h}$ and $\mathbf{r}$ are taken from $\mathbf{e}^{out}$ and $\mathbf{r}^{out}$.
Note that in fact any KGC decoder can be considered here, while this is not the focus of this paper. We leave the more explorations to the future work. 

Then we use binary cross entropy loss to measure the matching between $\mathbf{q}$ and potential answer entities $\mathbf{t}$: 
\begin{equation}
  \begin{aligned}
    \mathcal{L} = -\frac{1}{N} \sum_{t} \mathrm{1}(t) \cdot \log \left(m(\mathbf{q}, \mathbf{t})\right) + \\ 
    (1-\mathrm{1}(t)) \cdot \log (1-m(\mathbf{q}, \mathbf{t}))
  \end{aligned}
\end{equation}
where $N$ is the total number of candidate entities, and $m(\mathbf{q}, \mathbf{t}) \in [0, 1]$ is the matching function of query $\mathbf{q}$ and entity $\mathbf{t}$. We use dot product in this work: 
\begin{equation}
    m(\mathbf{q}, \mathbf{t}) = \mathrm{Sigmoid}(\mathbf{q}^{\mathrm{T}}\mathbf{t})
\end{equation}
$\mathrm{1}(t) \in \{0, 1\}$ is the denotation function that outputs $1$ for positive entity and $0$ for negative entity. 

\section{Experiments}

\subsection{Experiment Setup}
\label{sec: experiment setup}
We conduct experiments of Knowledge Graph Completion task on two commonly used public datasets: FB15k-237 \cite{2015_Toutanova_FB15k-237} and WN18RR \cite{AAAI_2018_Dettmers_ConvE_WN18RR}. The detailed introduction of two dataset are provided in appendix \ref{ap: dataset}.

We measure the model performance by five frequently used metrics: MRR (the Mean Reciprocal Rank of correct entities), MR (the Mean Rank of correct entities), Hits@1, Hits@3, Hits@10 (the accuracy of correct entities ranking in top 1/3/10). 
We follow the filtered setting protocol \cite{NeurIPS_2013_Bordes_TransE} for evaluation, i.e. all the other true entities appearing in train, valid and test set are excluded when ranking. 
In addition, based on the observation of \cite{ACL_2020_Sun_re-eval}, to eliminate the influence of abnormal score distribution, if prediction targets have the same score with multiple other entities, we take the average of upper bound and lower bound rank as the result.
Additional experimental details are provided in the appendix \ref{ap: experiment config}.

\subsection{Results of Knowledge Graph Completion task}
Our baselines are selected from three categories which are \textbf{Translational Distance Models}: TransE \cite{NeurIPS_2013_Bordes_TransE}, RotatE \cite{ICLR_2019_Sun_RotatE}, PaiRE \cite{ACL_2021_Chao_PaiRE}; \textbf{Semantic Matching Models}: DistMult \cite{ICLR_2015_Yang_DistMult}, ComplEx \cite{ICML_2016_Trouillon_ComplEx}, TuckER \cite{EMNLP_2019_Balazevic_TuckER}, ConvE \cite{AAAI_2018_Dettmers_ConvE_WN18RR}, InteractE \cite{AAAI_2020_Vashishth_InteractE}, \textsc{ProcrustEs} \cite{NAACL_2021_Peng_ProcrustEs}; \textbf{GNN-based Models}: R-GCN \cite{ESWC_2018_Schlichtkrull_R-GCN}, KBGAT \cite{ACL_2019_Nathani_KBGAT}, SACN \cite{AAAI_2019_Shang_SACN}, A2N \cite{ACL_2019_Bansal_A2N}, CompGCN \cite{ICLR_2020_Vashishth_CompGCN}. The results are demonstrated in table \ref{tab: kgc result}, from which we can know that: 
\begin{itemize}
    \item In view of the five metrics on two datasets, SE-GNN achieves 9 of 10 SOTAs, which is an
    overall best performance compared to baselines. 
    And for the exception of MR report on WN18RR, SE-GNN still gets the competitive result with regard to the most baselines. 
    \item SE-GNN obtains obvious improvement compared to CompGCN, which is a typical GNN-based KGE model. This shows that the aggregation function in SE-GNN for modeling three levels of SE information is a more sufficient way and performs a better extrapolation ability. 
    \item In addition, we can see that the improvement of SE-GNN is more evident on FB15k-237 dataset. We think this is because in FB15k-237 there are more than 200 types of relation (table \ref{tab: dataset} in appendix \ref{ap: dataset}) and the data interactions are very complex, which makes the extrapolation scenario more challenging. In this case, the explicit modeling of SEs will play a more important role to help extrapolation.
\end{itemize}
In the overall consideration across metrics on two datasets, SE-GNN obtains the best extrapolation performance on unseen test data, which indicates the effectiveness of our proposed method.

\begin{table*}
    \setlength\tabcolsep{5.7pt}
    \renewcommand\arraystretch{1.1}
    \centering
    \begin{tabular}{lccccccccccc}
      \toprule
      \multirow{2}{*}{\textbf{Models}} & \multicolumn{5}{c}{\textbf{FB15k-237}} & & \multicolumn{5}{c}{\textbf{WN18RR}} \\
      \cline{2-6} \cline{8-12}
      & MRR & MR & H@1 & H@3 & H@10 & & MRR & MR & H@1 & H@3 & H@10 \\
      \hline \hline
      \multicolumn{12}{l}{\textbf{Translational Distance}} \\
      TransE \cite{NeurIPS_2013_Bordes_TransE}$^{\dagger}$     & .330 & 173 & .231 & .369 & .528 & & .223 & 3380 & .014 & .401 & .529 \\
      RotatE \cite{ICLR_2019_Sun_RotatE}      & .338 & 177 & .241 & .375 & .533 & & .476 & 3340 & .428 & .492 & .571 \\
      PaiRE \cite{ACL_2021_Chao_PaiRE}        & .351 & 160 & .256 & .387 & .544 & & - & - & - & - & - \\
      \hline
      \multicolumn{12}{l}{\textbf{Semantic Matching}} \\ 
      DistMult \cite{ICLR_2015_Yang_DistMult}$^{\dagger}$      & .308 & 173 & .219 & .336 & .485 & & .439 & 4723 & .394 & .452 & .533 \\
      ComplEx  \cite{ICML_2016_Trouillon_ComplEx}$^{\dagger}$ & .323 & 165 & .229 & .353 & .513 & & .468 & 5542 & .427 & .485 & .554 \\
      TuckER\cite{EMNLP_2019_Balazevic_TuckER} & .358 & -   & .266 & .394 & .544 & & .470 & -    & .443 & .482 & .526 \\
      ConvE  \cite{AAAI_2018_Dettmers_ConvE_WN18RR} & .325 & 244 & .237 & .356 & .501 & & .430 & 4187 & .400 & .440 & .520\\
      InteractE \cite{AAAI_2020_Vashishth_InteractE} & .354 & 172 & .263 & - & .535 & & .463 & 5202 & .430 & - & .528\\ 
      \textsc{ProcrustEs} \cite{NAACL_2021_Peng_ProcrustEs} & .345 & - & .249 & .379 & .541 & & .474 & - & .421 & .502 & .569\\ 
      \hline
      \multicolumn{12}{l}{\textbf{GNN-based}} \\ 
      R-GCN \cite{ESWC_2018_Schlichtkrull_R-GCN}      & .248 & -   & .151 & - & .417 & & -  & -  & -  & - & -    \\
      KBGAT \cite{ACL_2019_Nathani_KBGAT}$^{\ddagger}$     & .157 & 270 & - & - & .331 & & .412 & \textbf{1921} & - & - & .554 \\
      SACN \cite{AAAI_2019_Shang_SACN}         & .350 & -   & .260 & .390 & .540 & & .470 & -    & .430 & .480 & .540\\
      A2N \cite{ACL_2019_Bansal_A2N}         & .317 & -   & .232 & .348 & .486 &  & .450 & -    & .420 & .460 & .510\\
      CompGCN\cite{ICLR_2020_Vashishth_CompGCN}       & .355 & 197 & .264 & .390 & .535 & & .479 & 3533 & .443 & .494 & .546 \\
      \hline
        \textbf{SE-GNN} (ours)        & \textbf{.365} & \textbf{157} & \textbf{.271} & \textbf{.399} & \textbf{.549} & & \textbf{.484} & 3211 & \textbf{.446} & \textbf{.509} & \textbf{.572}\\
      \bottomrule
    \end{tabular}
    \caption{
    Model reports on FB15k-237 and WN18RR test set. The best results are in bold. $^{\dagger}$ denotes that we reproduce the results using the code\footnotemark. $^{\ddagger}$ means that the results of KBGAT are from \cite{ACL_2020_Sun_re-eval} because original results suffer from same score evaluation problem, which is discussed in section \ref{sec: experiment setup}. 
    Other results are from the published paper.
    }
    \label{tab: kgc result}
\end{table*}

\subsection{Effective Modeling of Semantic Evidences}

\begin{figure}[t]
    \centering
    \includegraphics[width=\columnwidth]{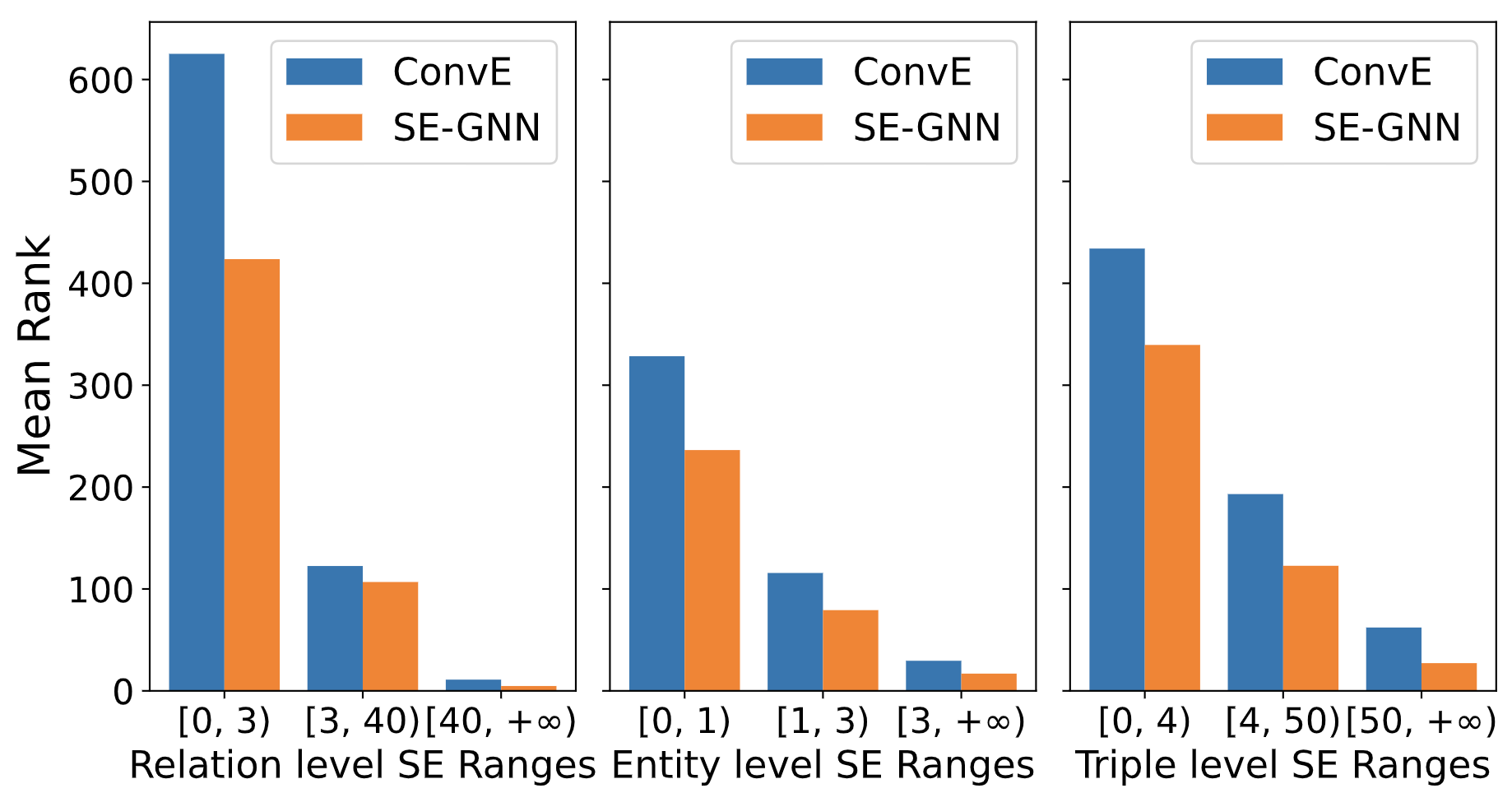}
    \caption{The performance comparation of SE-GNN and ConvE under the same Semantic Evidence range.}
    \label{fig: segnn_conve_se2rank}
\end{figure}

In this section, we tend to verify that SE-GNN is capable of effectively modeling the Semantic Evidences. Like in section \ref{sec: SE Verification}, we evaluate the extrapolation performance of SE-GNN in different SE ranges. To control the variables, we compare the results with ConvE, which is our selected decoder in SE-GNN. So the only difference here is that in SE-GNN, explicit modeling of three levels of Semantic Evidence is introduced before decoder (equation \ref{eq: SE-GNN output}), while in ConvE, entity and relation embedding are directly fed into the decoder, 
with implicit modeling of SE information. 
The results are demonstrated in figure \ref{fig: segnn_conve_se2rank}. We can see that SE-GNN performs better for all levels of SE across all ranges, which shows SE-GNN can capture the SE information more effectively and possess better extrapolation ability.

\subsection{Ablation Study of each Semantic Evidence}
\begin{table}
    \setlength\tabcolsep{4.8pt}
    \centering
    \begin{tabular}{lcccc}
      \toprule
      \multirow{2}{*}{\textbf{Models}} & \multicolumn{4}{c}{\textbf{FB15k-237}} \\
      \cline{2-5}
      & MRR & MR & H@1 & H@10 \\
      \hline \hline
      \textbf{SE-GNN}     & \textbf{.365} & \textbf{157} & \textbf{.271} & \textbf{.549} \\
      w/o relation SE & .361 & 168 & .264 & .542 \\
      w/o triple SE & .359 & 173 & .262 & .537 \\
      w/o entity SE & .360 & 172 & .265 & .539 \\
      w/o relation \& entity SE & .357 & 179 & .257 & .532 \\
      w/o relation \& triple SE & .355 & 181 & .254 & .535 \\
      w/o entity \& triple SE & .352 & 185 & .249 & .525 \\
      \bottomrule
    \end{tabular}
    \caption{Ablation study of three SEs, where w/o means removing the corresponding modeling part in SE-GNN.}
    \label{tab: ablation study}
\end{table}

To evaluate the effect of each SE part, we do the ablation study of only removing one SE modeling part and simultaneously removing two of them. The results are demonstrated in table \ref{tab: ablation study}. We can observe that the performance degrades for all six variants of SE-GNN, which shows the effectiveness of each SE modeling part. 

In addition, we consider that for most GNN-based KGE works like R-GCN \cite{ESWC_2018_Schlichtkrull_R-GCN}, CompGCN \cite{ICLR_2020_Vashishth_CompGCN}, the core idea is to merge the relation and entity together when neighbor aggregating. This can be regarded as the \textit{w/o relation \& entity SE} variant of SE-GNN, which only models the triple SE part. While both our SE extrapolation analysis and the ablation experiments show that it is insufficient, and separately modeling relation and entity information are also beneficial for KGE task. 

\footnotetext{\burl{https://github.com/DeepGraphLearning/KnowledgeGraphEmbedding}, commit ID: 2e440e0}

\section{Conclusion}
In this paper, we make the attempt to study the KGE extrapolation problem from a data relevant and model independent view. 
We show that there are three levels of Semantic Evidence that play an important role when predicting unseen data, which are the co-occurrence between relations and entities, the path connection between entities, and the similarity between observed entities and predicted entities.
Then we verify the effectiveness of SEs through extensive quantitative experiments and qualitative case study. Based on such observation, we design a novel SE-GNN model to obtain more extrapolative knowledge representation and achieve consistent improvement on different datasets. Some future directions include exploiting more extrapolative evidences and designing more elaborated SE modeling method. 

\begin{figure*}
    \centering 
    \includegraphics[width=\textwidth]{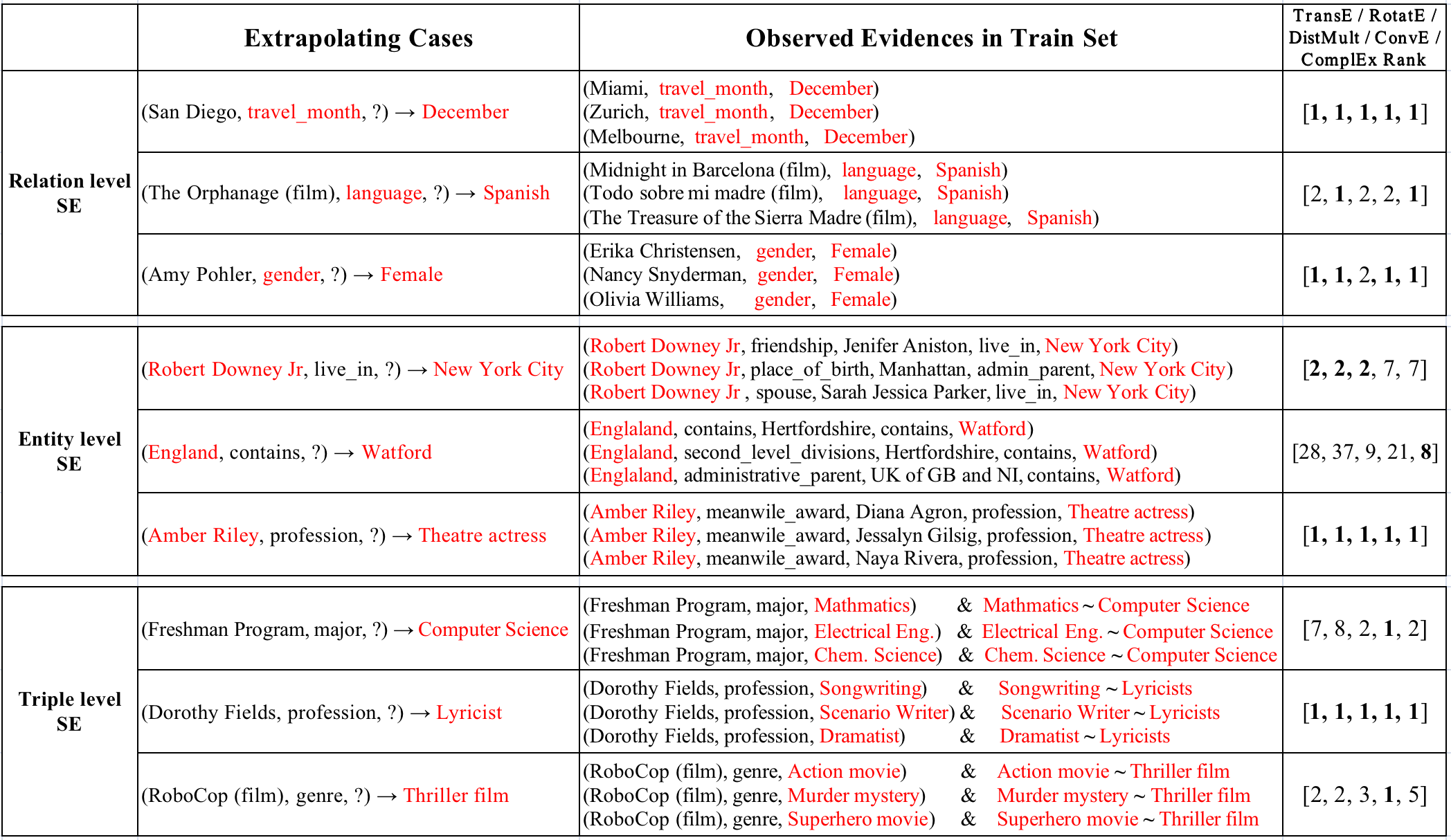} 
    \caption{Case study of Semantic Evidence on FB15k-237 dataset. The symbol $\sim$ means that two entities are semantically similar. The rank result is the prediction rank of correct entity in all entities. The best results are marked as bold.} 
    \label{fig: case study} 
\end{figure*}
\newpage

\appendix

\section*{Appendix}

\section{Case study of Semantic Evidence}
\label{ap: case study}

In this section, we tend to provide an intuitive demonstration about how the Semantic Evidence helps extrapolate to unseen data and hence this information is important for KGE task. For each Semantic Evidence, we select multiple example cases from FB15k-237 test set, and list their corresponding evidences and prediction results of different KGE models. The cases are demonstrated in figure \ref{fig: case study}. 

For relation SE, the idea is that the co-occurrence between relation and tail entity can help extrapolate. For example, for the extrapolative prediction \texttt{(San Diego, travel\_month, ?)} $\rightarrow$ \texttt{December}, if the model has observed large amounts of co-occurrences of \texttt{travel\_month} and \texttt{December} in train set, it will be aware of the month type of \texttt{December} and also know that \texttt{December} is a popular time for traveling, which is beneficial for the model to perform a new \texttt{travel\_month} prediction on \texttt{December} entity. Note that there are three levels of SE information that can help extrapolate in the meantime, and in this case relation evidence just servers as one part. 

For entity SE, it is that the connection or path between entities can help extrapolation. Like the case \texttt{(Robert Downey Jr, live\_in, ?)} $\rightarrow$ \texttt{New York City}, in the train set we know that \texttt{Downey} was born in \texttt{New York City}, his wife lives in \texttt{New York City}, his friend lives in \texttt{New York City}, etc. These connections between \texttt{Robert Downey Jr} and \texttt{New York City} will enhance their semantic relevancy and help the model to predict ``some relation'' between them, such as \texttt{live\_in}. 

For triple SE, it follows the idea that if prediction holds for one entity, it should also hold for a similar one. For example, if we have known that \texttt{Freshman Program} contains the \texttt{major} of \texttt{Mathematics}, \texttt{Electrical Engineering}, \texttt{Chemistry Science}, it is natural to infer that it also contains \texttt{Computer Science}, which is a similar major of the observed ones. 

Above three levels of SE are important to help the model do extrapolating. For a further illustration, we reproduce several typical KGE models and list their prediction in rightmost column. We can see that for these unseen cases with abundant SE information, all the models can perform a good extrapolative prediction, which verifies the effectiveness of the proposed Semantic Evidence concept.

\begin{figure*}
    \centering 
    \includegraphics[width=\textwidth]{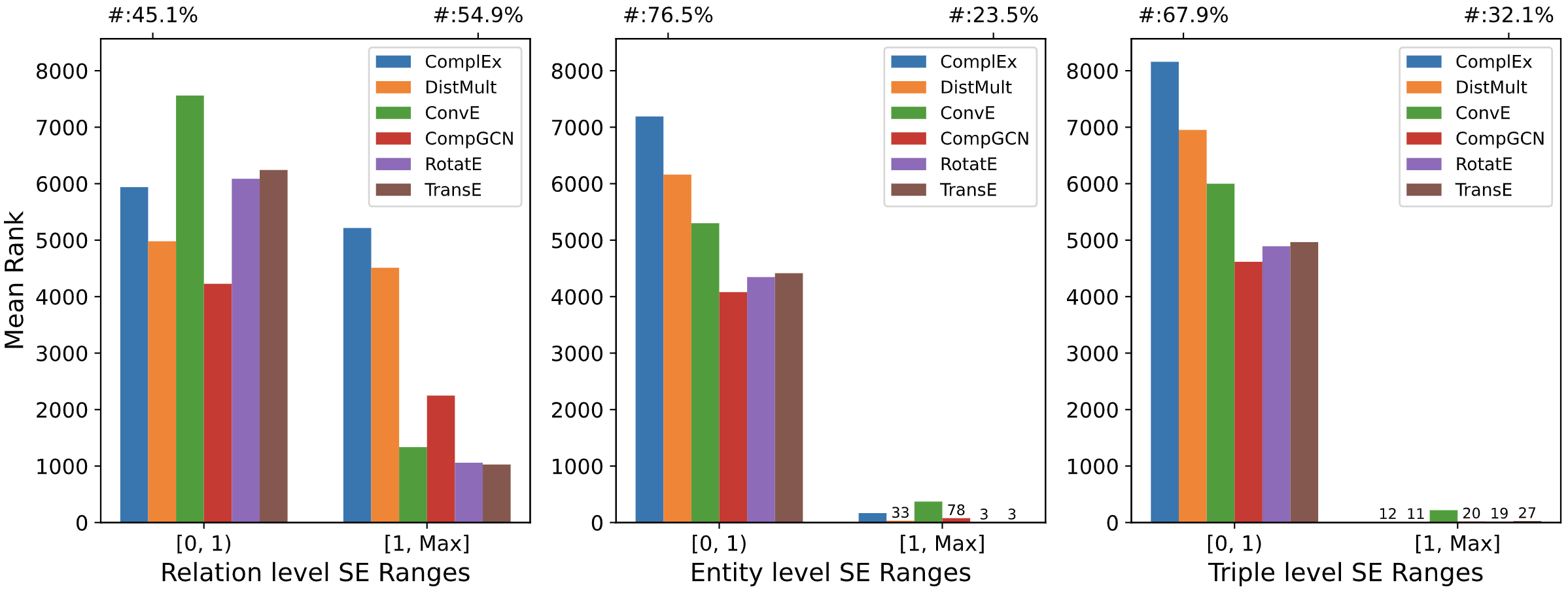} 
    \caption{Extrapolation performance of KGE models on different SE ranges on WN18RR dataset. The bottom x-axis denotes the range value, the top x-axis denotes the data portion of each range and y-axis denotes the Mean Rank of model prediction result (low value indicates good performance, and $1$ is the best).} 
    \label{fig: wn_baselines_se2rank}
\end{figure*}

\section{Correlation between SE and Extrapolation Performance on WN18RR Dataset}
\label{ap: wn_baselines_se2rank}
In this section, we compute the extrapolation performance of various KGE models on different SE ranges on WN18RR dataset. The results are demonstrated in figure \ref{fig: wn_baselines_se2rank}. 
Because there are much low evidence data in WN18RR dataset, it is hard to evenly divide the test data into three ranges like figure \ref{fig: fb_baselines_se2rank}. Therefore we divide the data into $[0, 1)$ and $[1, \mathrm{Max}]$ two ranges instead, which can respectively represent the data \textbf{with evidence} and \textbf{without evidence} ($\mathrm{Max}$ denotes the max value of corresponding evidence metric). 
From the results, we can see that the correlation between evidence strength and extrapolation performance also holds on WN18RR dataset, for all models and across all evidences, which further verifies the effectiveness of SE.
Note that because of the various data characteristics, different dataset may reveal a different focus on three levels of SE when extrapolating. 
Like in WN18RR dataset, there are only 11 types of relation (demonstrated in table \ref{tab: dataset}). Such simple pattern makes relation level SE a less important role compared to FB15k-237 dataset, which has an abundant relation set of 237 types. Hence in figure \ref{fig: wn_baselines_se2rank}, there is a weaker downward trend for relation level SE, and the entity and triple level SE play a more important role when extrapolating. 

\section{Dataset Statistics}
\label{ap: dataset}
In this section we provide the information of FB15k-237 and WN18RR dataset used in our experiment.
\begin{itemize}
    \item FB15k-237 \cite{2015_Toutanova_FB15k-237} contains entities and relations from Freebase, which is a large commonsense knowledge base. FB15k-237 is a pruned version of FB15k \cite{NeurIPS_2013_Bordes_TransE} dataset, with duplicate and inverse relations being removed to prevent direct prediction. Furthermore, FB15k-237 also ensures that every triple $(h, r, t)$ in valid and test set does not have any direct connection $(h, r', t)$ in train set, to make the prdiction more challenging. 
    \item WN18RR \cite{AAAI_2018_Dettmers_ConvE_WN18RR} is derived from WordNet, a lexical database of semantic relations between words. Similar to FB15k-237, WN18RR is pruned from WN18 \cite{NeurIPS_2013_Bordes_TransE} dataset by removing the duplicate and inverse relations, while there is no direct connection restriction in WN18RR.
\end{itemize}
Statistics of two datasets are summarized in table \ref{tab: dataset}.

\begin{table}[h]
    \centering
    \begin{tabular}{lcc}
      \toprule
      \textbf{Dataset} & \textbf{FB15k-237} & \textbf{WN18RR}\\
      \hline\hline
      \# entity          & 14,541     & 40,943  \\
      \# relation        & 237        & 11      \\ 
      \# train triple    & 272,115    & 86,835  \\
      \# valid triple    & 17,535     & 3,034   \\ 
      \# test triple     & 20,466     & 3,134   \\
      \bottomrule
    \end{tabular}
    \caption{Dataset statistics}
    \label{tab: dataset}
\end{table}

\section{Experimental Details}
\label{ap: experiment config}
In this section we discuss some more details of the experiment implementation. Following CompGCN \cite{ICLR_2020_Vashishth_CompGCN}, we transform the knowledge graph to undirected graph, by introducing an inverse edge $(t, r^{-1}, h)$ for each edge $(h, r, t)$, which aims to pass the information bidirectionally and enhance graph connectivity. In addition, like ConvE \cite{AAAI_2018_Dettmers_ConvE_WN18RR}, we also introduce an inverse version for each relation when predicting. For the two directions $(h, r, ?) \rightarrow t$ and $(?, r, t) \rightarrow h$ of a triple prediction, we transform them as $(h, r, ?) \rightarrow t$ and  $(t, r^{-1}, ?) \rightarrow h$, which can unify the prediction format and improve computational efficiency.

Furthermore, during the aggregation process of SE-GNN, for each training batch, we randomly remove a proportion of corresponding edges in the knowledge graph. This can prevent the information leakage problem, i.e. the model has seen the prediction edges when aggregating. This can also guide the model to learn the interactions between existed edges and prediction missing edges when aggregating, which is a closer scenario of extrapolation.

\clearpage

\section*{Acknowledgments}
This work was supported by the the Youth Innovation Promotion Association CAS (No.2018192), the National Natural Science Foundation of China (No. 62102421) , and Intelligent Social Governance Platform, Major Innovation \& Planning Interdisciplinary Platform for the "Double-First Class" Initiative, Renmin University of China. 

\bibliography{aaai22}

\end{document}